# Scalable $k$-Means Clustering via Lightweight Coresets

Olivier Bachem
Google Brain
ETH Zurich
bachem@google.com

Mario Lucic
Google Brain
lucic@google.com

Andreas Krause
ETH Zurich
krausea@ethz.ch

## ABSTRACT

*Coresets* are compact representations of data sets such that models trained on a coreset are *provably* competitive with models trained on the full data set. As such, they have been successfully used to scale up clustering models to massive data sets. While existing approaches generally only allow for multiplicative approximation errors, we propose a novel notion of *lightweight coresets* that allows for both multiplicative and additive errors. We provide a single algorithm to construct lightweight coresets for $k$-means clustering as well as soft and hard Bregman clustering. The algorithm is substantially faster than existing constructions, embarrassingly parallel, and the resulting coresets are smaller. We further show that the proposed approach naturally generalizes to statistical $k$-means clustering and that, compared to existing results, it can be used to compute smaller summaries for empirical risk minimization. In extensive experiments, we demonstrate that the proposed algorithm outperforms existing data summarization strategies in practice.

## 1 INTRODUCTION

A recent, unprecedented increase in the size of data sets has led to new challenges for machine learning as many traditional algorithms fail to scale to such *massive* data sets. For instance, many algorithms that have a superlinear computational complexity in the input size become computationally infeasible in the presence of millions or billions of data points. Similarly, many algorithms assume that the data can be accessed multiple times on a single machine. However, in a practical big data setting, one may only be able to see each data point a few times and the whole data set may be distributed across a cluster of machines. Traditional algorithms optimized for single machines and small data set sizes are not suitable for this setting.

*Coresets* are a proven data summarization approach that can be used to scale clustering problems to massive data sets. Coresets are small, weighted subsets of the original data set such that models trained on the coreset are *provably* competitive with models trained on the full data set. As such, they can be used to speed up inference while retaining strong theoretical guarantees on the solution quality. One first constructs a coreset — usually in linear time — and then uses any algorithm that works on weighted data to solve the clustering problem on the coreset. As the coreset size is usually sublinear in or even *independent* of the number of data points, computationally intensive inference algorithms with superlinear complexity may be applied on coresets.

Coresets have been successfully constructed for a wide variety of clustering problems. Yet, even linear-time coreset constructions can be hard to scale to massive data sets in a distributed setting. For example, the state-of-the-art coreset construction for $k$-means [21] requires $k$ sequential passes through the full data set. While there are approaches to construct coresets in a distributed fashion (discussed in Section 2), they usually increase both the total computational effort and the required coreset sizes substantially. Consequently, existing coreset constructions are not suitable for the practical setting where the data is distributed across a cluster and can only be processed once or twice in parallel.

**Our contributions.** We propose a novel approach to coreset construction for $k$-means clustering that retains the benefits of previous coreset constructions at a fraction of the cost, that works in the distributed setting and that can be applied naturally to the statistical $k$-means clustering problem. In particular, we

(i) introduce and motivate the novel notion of *lightweight* coresets that allows for both multiplicative and additive errors,
(ii) provide a simple and embarrassingly parallel algorithm to construct such lightweight coresets using only two full passes through the data set,
(iii) prove a sufficient coreset size of $O\left(\frac{dk \log k + \log \frac{1}{\delta}}{\epsilon^2}\right)$ which is only linear in the dimensionality $d$ and near-linear in the number of the clusters $k$,
(iv) extend the results to hard and soft clustering with a large class of Bregman divergences,
(v) show that the approach enables substantially smaller summaries for statistical $k$-means clustering, and
(vi) confirm the practical utility of the proposed method by comparing it to existing data summarization strategies in extensive experiments.

## 2 BACKGROUND AND RELATED WORK

**$k$-means clustering.** Let $\mathcal{X}$ denote a set of points in $\mathbb{R}^d$. The *$k$-means clustering problem* is to find a set $Q$ of $k$ cluster centers in $\mathbb{R}^d$ such that the quantization error $\phi_{\mathcal{X}}(Q)$ is minimized, where

$$\phi_{\mathcal{X}}(Q) = \sum_{x \in \mathcal{X}} \mathrm{d}(x, Q)^2 = \sum_{x \in \mathcal{X}} \min_{q \in Q} \|x - q\|_2^2.$$

For general $k$-clustering the squared Euclidean distance is replaced with the corresponding divergence measure d. For a weighted set $C$ with corresponding weights $w : C \to \mathbb{R}_{\geq 0}$, the quantization error is defined as $\phi_C(Q) = \sum_{x \in C} w(x) \, \mathrm{d}(x, Q)^2$. The quantization error of the optimal solution of $k$ centers is denoted by $\phi_{OPT}^k(\mathcal{X})$.

**Coresets.** A coreset is a weighted subset of the data such that the quality of any clustering evaluated on the coreset closely approximates the quality on the full data set. For the $k$-means clustering problem, this property is usually formalized as follows [4, 14, 21]: A weighted set $C$ is a $(\varepsilon, k)$-coreset for $\mathcal{X}$ if for any $Q \subset \mathbb{R}^d$ of cardinality at most $k$

$$|\phi_{\mathcal{X}}(Q) - \phi_C(Q)| \leq \varepsilon \phi_{\mathcal{X}}(Q). \tag{1}$$

This is a strong theoretical guarantee as the cost evaluated on the coreset $\phi_C(Q)$ has to approximate the cost on the full data set $\phi_{\mathcal{X}}(Q)$ up to a $1 \pm \varepsilon$ multiplicative factor *simultaneously* for all possible sets of cluster centers. As a direct consequence, solving on the coreset yields provably competitive solutions when evaluated on the full data set [21]. More formally, if $C$ is a $(\varepsilon, k)$-coreset of $\mathcal{X}$ with $\varepsilon \in (0, 1/3)$, it is shown that

$$\phi_{\mathcal{X}}\left(Q_C^*\right) \leq (1 + 3\varepsilon) \phi_{\mathcal{X}}\left(Q_{\mathcal{X}}^*\right) \tag{2}$$

where $Q_C^*$ denotes the optimal solution of $k$ centers on $C$ and $Q_{\mathcal{X}}^*$ denotes the optimal solution on $\mathcal{X}$. Any $\alpha$-approximation can be used on the $(\varepsilon, k)$-coreset to produce a $(1 + 3\varepsilon)\alpha$-approximation on the original data. As a result, any solver that works on weighted data may be used to solve the clustering problem on the coreset while retaining strong theoretical guarantees.

Coresets are usually small, *i.e.*, their size is logarithmic in or even independent of the number of data points. Hence, one may obtain a fast approximation of the optimal solution, even if the solver used on the coreset has superlinear complexity in the data set size.

**Coresets for $k$-means.** There is a rich history of coreset constructions for $k$-means [4, 7, 10, 13, 14, 19, 21]. The coreset construction by Lucic et al. [21] with both state-of-the-art theoretical and experimental results works as follows: In a first step, a rough approximation of the optimal solution is obtained using the seeding step of the well known k-means++ algorithm [1]. In a second step, this rough solution is used to bound the worst-case impact of each point on the objective function. In a third step, this is used to derive an importance sampling scheme that provides $(\varepsilon, k)$-coresets if enough points are sampled. Bachem et al. [4] show that the sufficient coreset size is bounded by $O\left(k \log k (dk \log k + \log \frac{1}{\delta}) \varepsilon^{-2}\right)$. However, there is a drawback to this approach: The coreset construction requires $k$ sequential passes through the data set. Constructing coresets for massive data sets can hence become prohibitively expensive even for moderate $k$ despite the linearity complexity in the data set size.

Balcan et al. [5] provide a low-communication distributed algorithm for $k$-means and $k$-median clustering on general network topologies and provide coresets for $k$-means of size $O(dk \log k / \varepsilon^4)$. In contrast, our algorithm computes coresets of size $O(dk \log k / \varepsilon^2)$.

**Other related work.** Previous constructions for $k$-means are based upon exponential grids [14] and building coresets along lines [13]. These geometrically inspired constructions suffer from coreset sizes exponential in the number of dimensions and are rarely practical. The use of sampling-based approaches was investigated by Chen [7] and Feldman et al. [11]. A general framework for constructing coresets was proposed by Langberg and Schulman [19] and Feldman and Langberg [10]. Feldman et al. [9] and Lucic et al. [22] applied this framework to construct coresets for the estimation of Gaussian mixture models. Recently, Bachem et al. [3] proposed a coreset construction for nonparametric clustering, Reddi et al. [24] for empirical risk minimization in supervised learning and Huggins et al. [16] for Bayesian logistic regression.

## 3 LIGHTWEIGHT CORESETS FOR $k$-MEANS

In this section, we propose and motivate *lightweight coresets* — a novel notion of coresets for $k$-means. In Section 6, we will then show how to extend the results to other divergence measures.

DEFINITION 1 (LIGHTWEIGHT CORESET FOR $k$-MEANS). *Let $\varepsilon > 0$ and $k \in \mathbb{N}$. Let $\mathcal{X} \subset \mathbb{R}^d$ be a set of points with mean $\mu(\mathcal{X})$. The weighted set $C$ is an $(\varepsilon, k)$-lightweight coreset of $\mathcal{X}$ if for any set $Q \subset \mathbb{R}^d$ of cardinality at most $k$*

$$|\phi_{\mathcal{X}}(Q) - \phi_C(Q)| \leq \frac{\varepsilon}{2} \phi_{\mathcal{X}}(Q) + \frac{\varepsilon}{2} \phi_{\mathcal{X}}(\{\mu(\mathcal{X})\}). \tag{3}$$

The notion of lightweight coresets may be interpreted as a relaxation of "traditional" coresets as defined in (1) that permits both an *additive* and *multiplicative* error. The $\frac{\varepsilon}{2} \phi_{\mathcal{X}}(Q)$ term allows the approximation error to scale with the quantization error and constitutes the "traditional" multiplicative part. The $\frac{\varepsilon}{2} \phi_{\mathcal{X}}(\{\mu(\mathcal{X})\})$ term scales with the variance of the data and corresponds to an additive approximation error that is invariant of the scale of the data.

We argue that the "additive" $\frac{\varepsilon}{2} \phi_{\mathcal{X}}(\{\mu(\mathcal{X})\})$ term is adequate for the following reason: In machine learning, one often tries to minimize the generalization error by performing empirical risk minimization on a finite sample. Yet, state-of-the art deviation bounds for finite samples only provide an additive error guarantee [2, 25]. Since one already incurs an additive error, one may thus accept the same additive error when optimizing on the finite sample. In fact, in Section 7, we will consider the statistical $k$-means clustering problem and will show that lightweight coresets may be used to generate small summaries suitable for empirical risk minimization.

The "multiplicative" $\frac{\varepsilon}{2} \phi_{\mathcal{X}}(Q)$ term is required for the following reason: Equation (3) needs to hold uniformly for all sets $Q$ of $k$ centers in $\mathbb{R}^d$. As such, with only the $\frac{\varepsilon}{2} \phi_{\mathcal{X}}(\{\mu(\mathcal{X})\})$ term, one would be able to construct the following adverse solution: If the cluster centers are placed arbitrarily far away from the data points, any difference on the left hand side in (3) would be arbitrarily large while the variance of the data would still bounded on the right hand side. Hence, without the multiplicative error term, there would exist no coreset $C$ satisfying (3) for all possible sets $Q$.

The primary motivation behind coresets is that the optimal solution obtained on the coreset is provably competitive with the optimal solution of the full dataset. For "traditional" coresets, this guarantee is multiplicative as defined in (2). In Theorem 1, we show that lightweight coresets directly imply a corresponding additive guarantee on the solution quality.

THEOREM 1. *Let $\varepsilon \in (0, 1]$. Let $X$ be any data set and $C$ be a $(\varepsilon, k)$-lightweight coreset of $X$, Denote by $Q_X^*$ an optimal $k$-means solution on $X$ and by $Q_C^*$ an optimal solution on $C$. Then, it holds that*
$$\phi_X\left(Q_C^*\right) \leq \phi_X\left(Q_X^*\right) + 4\varepsilon\phi_X(\{\mu(X)\}).$$

PROOF. By the lightweight coreset property, we have
$$\phi_C\left(Q_X^*\right) \leq \left(1 + \frac{\varepsilon}{2}\right)\phi_X\left(Q_X^*\right) + \frac{\varepsilon}{2}\phi_X(\{\mu(X)\}),$$
as well as
$$\phi_C\left(Q_C^*\right) \geq \left(1 - \frac{\varepsilon}{2}\right)\phi_X\left(Q_C^*\right) - \frac{\varepsilon}{2}\phi_X(\{\mu(X)\}).$$
Since by definition $\phi_C\left(Q_C^*\right) \leq \phi_C\left(Q_X^*\right)$ and $1 - \frac{\varepsilon}{2} \geq \frac{1}{2}$, it then holds that
$$\phi_X\left(Q_C^*\right) \leq \frac{1 + \frac{\varepsilon}{2}}{1 - \frac{\varepsilon}{2}}\phi_X\left(Q_X^*\right) + \frac{\varepsilon}{1 - \frac{\varepsilon}{2}}\phi_X(\{\mu(X)\})$$
$$\leq (1 + 2\varepsilon)\phi_X\left(Q_X^*\right) + 2\varepsilon\phi_X(\{\mu(X)\}).$$
The claim then follows since
$$\phi_X\left(Q_X^*\right) \leq \phi_X(\{\mu(X)\}).$$
□

Theorem 1 implies that, as we decrease $\varepsilon$, the true cost of the optimal solution obtained on the coreset approaches the true cost of the optimal solution on the full data set in an additive manner. In Section 4 we show that lightweight coresets allow us to obtain a more efficient coreset construction while retaining the empirical benefits of "traditional" coresets as evidenced in Section 8.

## 4 CONSTRUCTION OF LIGHTWEIGHT CORESETS

Our coreset construction is based on *importance sampling*. Let $q(x)$ be any probability distribution on $X$ and $Q$ any set of $k$ centers in $\mathbb{R}^d$. Then the quantization error may be rewritten as
$$\phi_X(Q) = \sum_{x \in X} q(x) \frac{d(x, Q)^2}{q(x)}.$$

The quantization error can hence be approximated by sampling $m$ points from $X$ using $q(x)$ and assigning them weights inversely proportional to $q(x)$. For any number of samples $m$ and any distribution $q(x)$, this yields an unbiased estimator of the quantization error. However, unbiasedness is not sufficient to guarantee a lightweight coreset as defined in Definition 1. In particular, (3) has to hold with probability $1 - \delta$ uniformly across all $k$-sized sets of centers $Q$. Obtaining such a stronger bound requires a suitable distribution $q(x)$ and a corresponding lower bound on the number of samples $m$. We suggest the following proposal distribution

$$q(x) = \underbrace{\frac{1}{2} \frac{1}{|X|}}_{(A)} + \underbrace{\frac{1}{2} \frac{d(x, \mu(X))^2}{\sum_{x' \in X} d(x', \mu(X))^2}}_{(B)}$$

**Algorithm 1** Lightweight coreset construction
---
**Require:** Set of data points $X$, coreset size $m$
1: $\mu \leftarrow$ mean of $X$
2: **for** $x \in X$ **do**
3: $\quad q(x) \leftarrow \frac{1}{2}\frac{1}{|X|} + \frac{1}{2}\frac{d(x,\mu)^2}{\sum_{x' \in X} d(x',\mu)^2}$
4: **end for**
5: $C \leftarrow$ sample $m$ weighted points from $X$ where each point $x$ has weight $\frac{1}{m \cdot q(x)}$ and is sampled with probability $q(x)$
6: **Return** lightweight coreset $C$
---

that has a natural interpretation as a mixture of two components. The first component (A) is the uniform distribution and ensures that all points are sampled with nonzero probability. The second component (B) samples points proportionally to their squared distance to the mean of the data. The intuition is that the points that are far from the mean of the data have a potentially large impact on the quantization error of a clustering. The component (B) ensures that these potentially important points are sampled frequently enough.

**Practical algorithm.** The resulting coreset construction is provided as pseudo code in Algorithm 1 and is extremely simple and practical: One calculates the mean of the data and then uses it to compute the importance sampling distribution $q(x)$. Finally, $m$ points are sampled with probability $q(x)$ from $X$ and assigned the weight $\frac{1}{m \cdot q(x)}$. The algorithm only requires two full passes through the data set resulting in a total computational complexity of $O(nd)$. There is no additional linear dependence on the number of clusters $k$ as in previous constructions [21] which is crucial in the setting where $k$ is even moderately large.

**Distributed implementation.** Algorithm 1 is embarrassingly easy to parallelize and can be implemented in a two-round distributed procedure: Let $X$ be partitioned across $p$ machines and let $X_i$ denote the points on the $i$-th machine. For each $x \in X$, let $x_j$ denote the $j$-th coordinate of $x$. We first compute the mean as follows. In the first round, each machine computes $|X_i|$, $U_{ij} = \sum_{x \in X_i} x_j$ and $V_{ij} = \sum_{x \in X_i} (x_j)^2$ and sends them back to the central machine. The central machine can then easily compute the global mean $\mu$, the quantization errors $\phi_{X_i}(\{\mu\})$ and the total quantization error $\phi_X(\{\mu\})$. For each machine $i$, we further keep track of the number of points $u_i$ to be sampled uniformly using (A) and the number of points $v_i$ to be sampled nonuniformly using (B). The central machine distributes the $m$ points to be sampled iteratively as follows: In each round, with probability $1/2$, it samples the machine $i$ with probability proportional to $|X_i|$ and increases the corresponding $u_i$ by one. Otherwise, it samples the machine $i$ with probability proportional to $\phi_{X_i}(\{\mu\})$ and increases $v_i$ by one. Each machine $i$ then obtains $\mu, u_i, v_i, \phi_{X_i}(\{\mu\})$ and $\phi_X(\{\mu\})$ and samples $u_i$ points uniformly at random and $v_i$ points proportionally to $d(x, \mu)^2$. For each sampled point, it further computes its weight and outputs the weight-point pair. Hence, there is no loss in approximation with respect to the single-machine implementation.

## 5 ANALYSIS

Our main result guarantees that Algorithm 1 computes lightweight coresets if sufficiently many points are sampled. The sufficient coreset size $m$ is independent of the number of data points, linear in the dimensionality and near-linear in the number of cluster centers.

**Theorem 2.** *Let $\varepsilon > 0, \delta > 0$ and $k \in \mathbb{N}$. Let $\mathcal{X}$ be a set of points in $\mathbb{R}^d$ and let $C$ be the output of Algorithm 1 with*

$$m \geq c \frac{dk \log k + \log \frac{1}{\delta}}{\varepsilon^2}$$

*where $c$ is an absolute constant. Then, with probability at least $1 - \delta$, the set $C$ is a $(\varepsilon, k)$-lightweight coreset of $\mathcal{X}$.*

**Proof.** We first derive an importance sampling distribution over $x \in \mathcal{X}$. Then, we show that by sampling a sufficient number of points from this importance sampling distribution one obtains a $(\varepsilon, k)$-lightweight coreset of $\mathcal{X}$. We first bound the importance of each data point $x \in \mathcal{X}$. For this, we define

$$\tilde{f}(Q) = \frac{1}{2|\mathcal{X}|} \phi_{\mathcal{X}}(Q) + \frac{1}{2|\mathcal{X}|} \phi_{\mathcal{X}}(\{\mu(\mathcal{X})\}) \quad (4)$$

where $\mu(\mathcal{X})$ denotes the mean of $\mathcal{X}$ and prove the following Lemma.

**Lemma 1.** *Let $\mathcal{X}$ be a set of points in $\mathbb{R}^d$ with mean $\mu(\mathcal{X})$. For all $x \in \mathcal{X}$ and $Q \subset \mathbb{R}^d$, it holds that*

$$\frac{d(x, Q)^2}{\tilde{f}(Q)} \leq \frac{16 \, d(x, \mu(\mathcal{X}))^2}{\frac{1}{|\mathcal{X}|} \sum_{x' \in \mathcal{X}} d(x', \mu(\mathcal{X}))^2} + 16. \quad (5)$$

**Proof.** By the triangle inequality and since

$$(|a| + |b|)^2 \leq 2a^2 + 2b^2,$$

we have for any $x \in \mathcal{X}$ and any $Q \subset \mathbb{R}^d$ that

$$d(\mu(\mathcal{X}), Q)^2 \leq 2 \, d(x, \mu(\mathcal{X}))^2 + 2 \, d(x, Q)^2.$$

Averaging across all $x \in \mathcal{X}$, we obtain

$$d(\mu(\mathcal{X}), Q)^2 \leq \frac{2}{|\mathcal{X}|} \sum_{x \in \mathcal{X}} d(x, \mu(\mathcal{X}))^2 + \frac{2}{|\mathcal{X}|} \sum_{x \in \mathcal{X}} d(x, Q)^2$$

$$= \frac{2}{|\mathcal{X}|} \phi_{\mathcal{X}}(\{\mu(\mathcal{X})\}) + \frac{2}{|\mathcal{X}|} \phi_{\mathcal{X}}(Q).$$

This implies that for all $x \in \mathcal{X}$ and $Q \subset \mathbb{R}^d$

$$d(x, Q)^2 \leq 2 \, d(x, \mu(\mathcal{X}))^2 + 2 \, d(\mu(\mathcal{X}), Q)^2$$

$$\leq 2 \, d(x, \mu(\mathcal{X}))^2 + \frac{4}{|\mathcal{X}|} \phi_{\mathcal{X}}(\{\mu(\mathcal{X})\}) + \frac{4}{|\mathcal{X}|} \phi_{\mathcal{X}}(Q).$$

We divide by $\tilde{f}(Q)$ as defined in (4) and obtain that

$$\frac{d(x, Q)^2}{\tilde{f}(Q)} \leq \frac{2 \, d(x, \mu(\mathcal{X}))^2 + \frac{4}{|\mathcal{X}|} \phi_{\mathcal{X}}(\{\mu(\mathcal{X})\}) + \frac{4}{|\mathcal{X}|} \phi_{\mathcal{X}}(Q)}{\frac{1}{2|\mathcal{X}|} \phi_{\mathcal{X}}(Q) + \frac{1}{2|\mathcal{X}|} \phi_{\mathcal{X}}(\{\mu(\mathcal{X})\})}$$

$$\leq \frac{2 \, d(x, \mu(\mathcal{X}))^2 + \frac{4}{|\mathcal{X}|} \phi_{\mathcal{X}}(\{\mu(\mathcal{X})\})}{\frac{1}{2|\mathcal{X}|} \phi_{\mathcal{X}}(\{\mu(\mathcal{X})\})} + \frac{\frac{4}{|\mathcal{X}|} \phi_{\mathcal{X}}(Q)}{\frac{1}{2|\mathcal{X}|} \phi_{\mathcal{X}}(Q)}$$

$$\leq \frac{16 \, d(x, \mu(\mathcal{X}))^2}{\frac{1}{|\mathcal{X}|} \sum_{x' \in \mathcal{X}} d(x', \mu(\mathcal{X}))^2} + 16$$

for all $x \in \mathcal{X}$ and $Q \subset \mathbb{R}^d$ which proves the Lemma. □

Lemma 1 implies that the ratio between the cost contribution of a single point $x \in \mathcal{X}$ and $\tilde{f}(Q)$ is bounded for all $Q \subset \mathbb{R}^d$ by

$$s(x) = \frac{16 \, d(x, \mu(\mathcal{X}))^2}{\frac{1}{|\mathcal{X}|} \sum_{x' \in \mathcal{X}} d(x', \mu(\mathcal{X}))^2} + 16.$$

We define $S = \frac{1}{|\mathcal{X}|} \sum_{x' \in \mathcal{X}} s(x')$ and note that $S = 32$ for any data set $\mathcal{X}$. The importance sampling distribution $q(x)$ in Algorithm 1 may hence be rewritten as

$$q(x) = \frac{1}{2} \frac{1}{|\mathcal{X}|} + \frac{1}{2} \frac{d(x, \mu(\mathcal{X}))^2}{\sum_{x' \in \mathcal{X}} d(x', \mu(\mathcal{X}))^2} = \frac{s(x)}{S|\mathcal{X}|}$$

for all $x \in \mathcal{X}$.

Consider the function

$$g_Q(x) = \frac{d(x, Q)^2}{\tilde{f}(Q) s(x)}$$

for all $x \in \mathcal{X}$ and $Q \subset \mathbb{R}^d$. Then, it holds for any $Q \subset \mathbb{R}^d$ that

$$\phi_{\mathcal{X}}(Q) = \sum_{x \in \mathcal{X}} d(x, Q)^2 = S|\mathcal{X}|\tilde{f}(Q) \sum_{x \in \mathcal{X}} \underbrace{\frac{s(x)}{S|\mathcal{X}|}}_{q(x)} \underbrace{\frac{d(x, Q)^2}{\tilde{f}(Q) s(x)}}_{g_Q(x)}$$

$$= S|\mathcal{X}|\tilde{f}(Q) \sum_{x \in \mathcal{X}} q(x) g_Q(x).$$

Defining the notation $\mathbb{E}_q[g_Q(x)] = \sum_{x \in \mathcal{X}} q(x) g_Q(x)$, we thus have

$$\phi_{\mathcal{X}}(Q) = 32|\mathcal{X}|\tilde{f}(Q) \mathbb{E}_q[g_Q(x)]. \quad (6)$$

As is standard in recent coreset constructions [4, 5, 10, 21], we apply the following seminal result of Li et al. [20] to estimate $\mathbb{E}_q[g_Q(x)]$ using a random sample drawn from the importance sampling distribution $q(x)$.

**Definition 2 (Haussler [15], Li et al. [20]).** *Fix a countably infinite domain $X$. The pseudo-dimension of a set $\mathcal{F}$ of functions from $X$ to $[0, 1]$, denoted by $\text{Pdim}(\mathcal{F})$, is the largest $d'$ such there is a sequence $x_1, \ldots, x_{d'}$ of domain elements from $X$ and a sequence $r_1, \ldots, r_{d'}$ of reals such that for each $b_1, \ldots, b_{d'} \in \{above, below\}$, there is an $f \in \mathcal{F}$ such that for all $i = 1, \ldots, d'$, we have $f(x_i) \geq r_i \iff b_i = above$.*

**Theorem 3 (Li et al. [20]).** *Let $\alpha > 0$, $\nu > 0$ and $\delta > 0$. Fix a countably infinite domain $X$ and let $P$ be any probability distribution over $X$. Let $\mathcal{F}$ be a set of functions from $X$ to $[0, 1]$ with $\text{Pdim}(\mathcal{F}) = d'$. Denote by $C$ a sample of $m$ points from $X$ independently drawn according to $P$ with*

$$m \geq \frac{c}{\alpha^2 \nu} \left( d' \log \frac{1}{\nu} + \log \frac{1}{\delta} \right)$$

*where $c$ is an absolute constant. Then, it holds with probability at least $1 - \delta$ that*

$$d_\nu \left( \mathbb{E}_P[f], \frac{1}{|C|} \sum_{x \in C} f(x) \right) \leq \alpha \quad \forall f \in \mathcal{F}$$

*where $d_\nu(a, b) = \frac{|a-b|}{a+b+\nu}$. Over all choices of $\mathcal{F}$ with $\text{Pdim}(\mathcal{F}) = d$, this bound on $m$ is tight.*

To obtain a uniform guarantee over all possible sets of cluster centers $Q$, we would like to apply Theorem 3 to approximate $\mathbb{E}_q[g_Q(x)]$ uniformly for the function family

$$\mathcal{G} = \left\{ g_Q(x) : Q \subset \mathbb{R}^d, |Q| \leq k \right\}.$$

For this, we require a bound on the pseudo-dimension of $\mathcal{G}$ which measures the richness of the function family and may be viewed as a generalization of the *Vapnik-Chervonenkis (VC) dimension*. Previous work [5, 10, 21] has used the property that the pseudo-dimension of $k$-means (or equivalently the function family $\mathcal{G}$) is essentially bounded by the VC dimension of $k$-fold intersections of half-spaces in $O(d)$-dimensional Euclidean space.[1]

However, Feldman and Langberg [10], Balcan et al. [5] as well as Lucic et al. [21] all use a different definition of pseudo-dimension than the underlying theorem by Li et al. [20]: they define it as the smallest integer $d$ that bounds the number of dichotomies induced by $\mathcal{G}$ on a sample of $m$ points by $m^d$. Such a bound on the growth function of the number of dichotomies may be regarded as a generalization of the *primal shattering dimension* in classical VC theory [12, 23]. Hence, the aforementioned papers bound the *primal shattering dimension* and not the *pseudo-dimension* of $k$-means by $O(dk)$, for example in Theorem 6 of Lucic et al. [21], and then proceed to (incorrectly) apply Theorem 3.

While both notions are closely related, one has to be careful: A primal shattering dimension of $d'$ only implies a VC dimension of at most $O(d' \log d')$ [12]. For the function family $\mathcal{G}$, this only provides a bound on the pseudo-dimension of $O(dk \log dk)$ and not $O(dk)$. In fact, it not known whether for $d > 3$ $k$-fold intersections of half-spaces are *VC-linear*, i.e., whether their VC dimension is bounded by $O(dk)$ [17].

In contrast, Lemma 1 of Bachem et al. [2] provides a sharper bound on the pseudo-dimension of $\mathcal{G}$ than the $O(dk \log dk)$ bound obtained via the primal shattering dimension. In particular, one obtains

$$\text{Pdim}(\mathcal{G}) \in O(dk \log k).$$

by setting $P$ in Lemma 1 of Bachem et al. [2] to be the empirical distribution associated with $\mathcal{X}$. Applying this bound to the coreset constructions in Feldman and Langberg [10], Balcan et al. [5], and Lucic et al. [21] leads to an additional $\log k$ (and not $\log dk$) factor in the coreset sizes presented in these papers.

We use Theorem 3 with this bound on the pseudo-dimension of $\mathcal{G}$ to approximate $\mathbb{E}_q[g_Q(x)]$ in (6). Choose $\alpha = \epsilon/96$ and $\nu = 1/2$ and note that the function $g_Q(x)$ is bounded in $[0, 1]$ for all $x \in \mathcal{X}$ and $Q \subset \mathbb{R}^d$ of cardinality at most $k$. By assumption, we consider the case

$$m \geq c \frac{dk \log k + \log \frac{1}{\delta}}{\epsilon^2},$$

where $c$ is an absolute constant. Hence, Theorem 3 implies that with probability at least $1 - \delta$

$$d_\nu \left( \mathbb{E}_q[g_Q(x)], \frac{1}{|C|} \sum_{x \in C} g_Q(x) \right) \leq \frac{\epsilon}{96}$$

uniformly for all sets $Q \subset \mathbb{R}^d$ of cardinality at most $k$.

---
[1]This can be shown by the use of a lifting map as in Theorem 6 of Lucic et al. [21].

The denominator in $d_\nu(\cdot, \cdot)$ is bounded by 3 since both arguments to $d_\nu(\cdot, \cdot)$ are bounded in $[0, 1]$. Hence, we have that

$$\left| \mathbb{E}_q[g_Q(x)] - \frac{1}{|C|} \sum_{x \in C} g_Q(x) \right| \leq \frac{\epsilon}{32}$$

for all sets $Q \subset \mathbb{R}^d$ of cardinality at most $k$. By multiplying both sides by $32|\mathcal{X}|\tilde{f}(Q)$ we obtain

$$\left| 32|\mathcal{X}|\tilde{f}(Q) \mathbb{E}_q[g_Q(x)] - \frac{32|\mathcal{X}|\tilde{f}(Q)}{|C|} \sum_{x \in C} g_Q(x) \right| \leq \epsilon |\mathcal{X}| \tilde{f}(Q)$$

for all sets $Q \subset \mathbb{R}^d$ of cardinality at most $k$. Let $(C, u)$ be a weighted set that contains all $x \in C$ with weight $u(x) = \frac{1}{|C|q(x)}$. From the definition of $g_Q(x)$ we have

$$\frac{32|\mathcal{X}|\tilde{f}(Q)}{|C|} \sum_{x \in C} g_Q(x) = \sum_{x \in C} \frac{1}{|C|q(x)} d(x, Q)^2$$

$$= \sum_{x \in C} u(x) d(x, Q)^2 = \phi_C(Q).$$

By (6) and the definition of $\tilde{f}(Q)$ in (4), we directly obtain the desired lightweight coreset property, *i.e.*, we have

$$|\phi_\mathcal{X}(Q) - \phi_C(Q)| \leq \frac{\epsilon}{2} \phi_\mathcal{X}(Q) + \frac{\epsilon}{2} \phi_\mathcal{X}(\{\mu(\mathcal{X})\})$$

for all sets $Q \subset \mathbb{R}^d$ of cardinality at most $k$ which concludes the proof of Theorem 2. □

## 6 EXTENSION TO $\mu$-SIMILAR BREGMAN DIVERGENCES

Building on the results of Lucic et al. [21], our results can be extended to both hard and soft clustering with $\mu$-similar Bregman divergences. This broad class of divergence measures includes among others the squared Mahalanobis distance, the KL-divergence and the Itakura-Saito distance.

Let $d_\phi(\cdot, \cdot)$ denote a $\mu$-similar Bregman divergence and $d_A(\cdot, \cdot)$ the corresponding squared Mahalanobis distance implied by $\mu$-similarity. The lightweight coreset property of Definition 1 can then be modified as follows. For hard clustering, we simply replace all occurrences of the squared Euclidean distance $d(\cdot, \cdot)^2$ by the Bregman divergence $d_\phi(\cdot, \cdot)$. For soft clustering, the *hard-min* is further replaced by a *soft-min*. In Algorithm 1, the squared Euclidean distance between the points and the mean of the data set is replaced by the squared Mahalanobis distance $d_A(\cdot, \cdot)$. The sufficient coreset size is $O\left(\frac{dk \log k + \log \frac{1}{\delta}}{\mu^2 \epsilon^2}\right)$ for Bregman hard clustering and $O\left(\frac{d^2 k^2 + \log \frac{1}{\delta}}{\mu^2 \epsilon^2}\right)$ for Bregman soft clustering.

The analysis in Section 5 should be adapted as follows: For hard clustering, an additional factor of $\frac{1}{\mu}$ needs to be added to the right hand side of Lemma 1 due to $\mu$-similarity, and $S$ should be scaled accordingly. For soft clustering, one further needs to apply Lemma 3 of Lucic et al. [21] to account for the *soft-min*. The corresponding pseudo dimension is bounded by $O(dk \log k)$ for Bregman hard and by $O(d^2 k^2)$ for Bregman soft clustering [21].

# 7 SMALL SUMMARIES FOR STATISTICAL $k$-MEANS CLUSTERING

In this section, we show how the notion of lightweight coresets has a natural application in generating small summaries for the *statistical $k$-means* problem. Up to now, we have considered the *empirical $k$-means clustering* problem where the goal is to optimize the quantization error

$$\phi_{\mathcal{X}}(Q) = \sum_{x \in \mathcal{X}} d(x, Q)^2$$

on a fixed data set $\mathcal{X}$. However, in many machine learning settings, one is often interested in minimizing the loss on an *unseen* data point $x$. The *statistical $k$-means* problem formalizes this setting as follows: We assume that $x$ is sampled from an underlying data generating distribution $P$ which we cannot observe directly, but can only draw independent samples from. The goal of the *statistical $k$-means* problem is to find to find $k$ cluster centers in $\mathbb{R}^d$ such that the *expected* quantization error

$$\mathbb{E}_P\left[d(x, Q)^2\right] = \int d(x, Q)^2 dP$$

is minimized. In the 1-means case, the optimal quantizer is the mean $\mu(P) = \int x dP$ and the optimal expected quantization error is equal to the variance $\sigma^2(P) = \mathbb{E}_P\left[d(x, \mu(P))^2\right]$.

The standard approach to find good clusterings in the statistical setting is the principle of *empirical risk minimization*. As the underlying distribution $P$ cannot be observed, one draws $n$ independent samples $\mathcal{X}$ from it instead. Then, one finds a solution that minimizes the empirical quantization error on the sample $\mathcal{X}$. The key intuition is that by the law of large numbers the expected quantization error $\mathbb{E}_P\left[d(x, Q)^2\right]$ is well approximated by the (average) empirical quantization error $\frac{1}{n}\phi_{\mathcal{X}}(Q)$ as $n \to \infty$.

Finite sample bounds quantify how large the sample $\mathcal{X}$ needs to be to achieve a fixed approximation error $\varepsilon$. Bachem et al. [2] show that it is sufficient to sample

$$n \in O\left(\frac{K(P)}{\varepsilon^2 \delta}\left(dk \log k + \log \frac{1}{\delta}\right)\right) \quad (7)$$

points to guarantee an approximation error similar to the one introduced in Section 3 where the kurtosis $K(P)$ is the normalized fourth moment of $P$, i.e.,

$$K(P) = \frac{\mathbb{E}_P\left[d(x, \mu(P))^4\right]}{\mathbb{E}_P\left[d(x, \mu(P))^2\right]^2}.$$

With an additional factor of $\frac{K(P)}{\delta}$, the sufficient sample size in (7) is substantially larger than the sufficient lightweight coreset size of $O\left(\frac{dk \log k + \log \frac{1}{\delta}}{\varepsilon^2}\right)$ in Theorem 2. Crucially, however, the sufficient sample size in (7) depends on the specific statistical clustering problem via the kurtosis $K(P)$ which depends on $P$. In fact, Bachem et al. [2] show that such a dependence on $P$ is necessary as for any $n$ there exists a distribution $P$ such that with high probability the quantization error on a sample of size $n$ provides a bad approximation of the expected quantization error.

The large sufficient sample size in (7) raises the question whether our approach to generate lightweight coresets may be used to obtain smaller summaries for the statistical $k$-means problem. We affirmatively answer this question with the following theorem.

THEOREM 4. *Let $\varepsilon \in (0, 1]$, $\delta \in (0, 1]$ and $k \in \mathbb{N}$. Let $P$ be a distribution on $\mathbb{R}^d$ with with finite kurtosis $K(P) < \infty$. Let $\mathcal{X}$ be a set of $n \in \Theta\left(\frac{K(P)}{\varepsilon^2 \delta}\left(dk \log k + \log \frac{1}{\delta}\right)\right)$ independent samples from $P$ and $C$ be the output of Algorithm 1 applied to $\mathcal{X}$ with $m \in \Theta\left(\frac{dk \log k + \log \frac{1}{\delta}}{\varepsilon^2}\right)$. Then, with probability at least $1 - \delta$, it holds for all $Q \subset \mathbb{R}^d$ of cardinality at most $k$ that*

$$\left|\mathbb{E}_P\left[d(x, Q)^2\right] - \frac{1}{n}\phi_C(Q)\right| \leq \frac{\varepsilon}{2}\mathbb{E}_P\left[d(x, Q)^2\right] + \frac{\varepsilon}{2}\sigma^2(P). \quad (8)$$

*The computational complexity is $\Theta\left(\frac{K(P)d}{\varepsilon^2 \delta}\left(dk \log k + \log \frac{1}{\delta}\right)\right)$.*

The key implication of Theorem 4 is that Algorithm 1 may be used to further summarize any sample $\mathcal{X}$ in the context of empirical risk minimization. While the sample $\mathcal{X}$ has to be sufficiently large to guarantee a good approximation quality, the size of the lightweight coreset $C$ is independent of $P$ and substantially smaller than the uniform sample $\mathcal{X}$.

The guarantee of the approximation quality in (8) is very similar to the guarantee in Definition 1 with both a multiplicative and additive error term based on the variance. The difference is that the (average) empirical quantization error $\frac{1}{n}\phi_{\mathcal{X}}(Q)$ is replaced by the expected quantization error $\mathbb{E}_P\left[d(x, Q)^2\right]$ while the empirical variance $\frac{1}{n}\phi_{\mathcal{X}}(\{\mu(X)\})$ is replaced by the variance of the underlying distribution $P$, i.e., $\sigma^2(P)$. The implications of (8) follow analogously to Theorem 1: The optimal solution on the lightweight coreset $C$ is provably competitive with the optimal quantizer of the underlying data generating distribution $P$.

On a practical note, Theorem 4 provides a simple yet effective strategy for empirical risk minimization for $k$-means clustering: Firstly, we sample as many points from the underlying distribution as possible, even if we cannot solve the full $k$-means clustering problem on this large sample. Secondly, we use Algorithm 1 to create a smaller summary for which one can solve the $k$-means clustering problem efficiently. In contrast to directly using a smaller sample from $P$, this approach allows us to retain strong theoretical guarantees as it may consider samples from $P$ that are very important for the clustering problem but are only infrequently sampled according to $P$ and hence would be missed by a small uniform sample.

PROOF OF THEOREM 4. We instantiate Theorem 1 of Bachem et al. [2] with $\varepsilon' = \frac{\varepsilon}{3}$ and $\delta' = \frac{\delta}{2}$. Then, with probability at least $1 - \frac{\delta}{2}$, it holds for all $Q \subset \mathbb{R}^d$ of cardinality at most $k$ that

$$\left|\mathbb{E}_P\left[d(x, Q)^2\right] - \frac{1}{n}\phi_{\mathcal{X}}(Q)\right| \leq \frac{\varepsilon}{6}\mathbb{E}_P\left[d(x, Q)^2\right] + \frac{\varepsilon}{6}\sigma^2(P). \quad (9)$$

Similarly, we consider Theorem 2 with $\varepsilon' = \frac{\varepsilon}{3}$ and $\delta' = \frac{\delta}{2}$. Together with Definition 1, this implies that with probability at least $1 - \frac{\delta}{2}$, we have for all $Q \subset \mathbb{R}^d$ of cardinality at most $k$ that

$$|\phi_{\mathcal{X}}(Q) - \phi_C(Q)| \leq \frac{\varepsilon}{6}\phi_{\mathcal{X}}(Q) + \frac{\varepsilon}{6}\phi_{\mathcal{X}}(\{\mu(\mathcal{X})\}). \quad (10)$$

By the union bound, this implies that both (9) and (10) hold with probability at least $1 - \delta$. As we only need to show that (8) in Theorem 4 holds with probability at least $1 - \delta$, we assume both (9) and (10) to hold for the remainder of the proof and show that this implies (8) in Theorem 4.

Since $\varepsilon \leq 1$, (9) implies that for all $Q \subset \mathbb{R}^d$ of cardinality at most $k$

$$\frac{1}{n}\phi_X(Q) \leq \left(1 + \frac{\varepsilon}{6}\right)\mathbb{E}_P\left[d(x,Q)^2\right] + \frac{\varepsilon}{6}\sigma^2(P)$$
$$\leq \frac{7}{6}\mathbb{E}_P\left[d(x,Q)^2\right] + \frac{1}{6}\sigma^2(P). \quad (11)$$

Similarly, since the mean is the optimal 1-quantizer, we have

$$\frac{1}{n}\phi_X(\{\mu(X)\}) \leq \frac{1}{n}\phi_X(\{\mu(P)\}) \leq \frac{4}{3}\sigma^2(P). \quad (12)$$

For all $Q \subset \mathbb{R}^d$ of cardinality at most $k$, let

$$\Delta(Q) = \left|\mathbb{E}_P\left[d(x,Q)^2\right] - \frac{1}{n}\phi_C(Q)\right|.$$

It holds that

$$\Delta(Q) \leq \underbrace{\left|\mathbb{E}_P\left[d(x,Q)^2\right] - \frac{1}{n}\phi_X(Q)\right|}_{(*)} + \underbrace{\left|\frac{1}{n}\phi_X(Q) - \frac{1}{n}\phi_C(Q)\right|}_{(**)}. \quad (13)$$

Using (9), we have that

$$(*) \leq \frac{\varepsilon}{6}\mathbb{E}_P\left[d(x,Q)^2\right] + \frac{\varepsilon}{6}\sigma^2(P). \quad (14)$$

Similarly, (9), (11) and (12) imply that

$$(**) \leq \frac{\varepsilon}{6}\frac{1}{n}\phi_X(Q) + \frac{\varepsilon}{6}\frac{1}{n}\phi_X(\{\mu(X)\})$$
$$\leq \frac{\varepsilon}{6}\frac{7}{6}\mathbb{E}_P\left[d(x,Q)^2\right] + \frac{\varepsilon}{6}\frac{1}{6}\sigma^2(P) + \frac{\varepsilon}{6}\frac{4}{3}\sigma^2(P) \quad (15)$$
$$= \frac{7\varepsilon}{36}\mathbb{E}_P\left[d(x,Q)^2\right] + \frac{9\varepsilon}{36}\sigma^2(P).$$

Combining (13), (14) and (15) yields

$$\Delta(Q) \leq \frac{13\varepsilon}{36}\mathbb{E}_P\left[d(x,Q)^2\right] + \frac{15\varepsilon}{36}\sigma^2(P)$$
$$\leq \frac{\varepsilon}{2}\mathbb{E}_P\left[d(x,Q)^2\right] + \frac{\varepsilon}{2}\sigma^2(P) \quad (16)$$

which proves (8). The computational complexity follows directly from Theorem 2. □

## 8 EXPERIMENTAL RESULTS

**Experimental setup.** We compare the lightweight coresets constructed with Algorithm 1 (denoted by LWCS) with two different subsampling methods: UNIFORM, the "naive" strategy of uniformly subsampling the data points, and CS, the state-of-the-art coreset construction by Lucic et al. [21].

For each of these methods, we generate subsamples of size $m \in \{1000, 2000, 5000, 10\,000, 20\,000\}$. We then use the state-of-the-art algorithm k-means++ [1] to solve the clustering problem on the subsample. We measure the elapsed time and then evaluate the clustering by computing the quantization error on the full data set. In addition, we run k-means++ on the full data set (denoted by FULL) and again measure the time and the solution quality as evaluated on the full data set.

We then compute the relative error $\eta$ for each method and subsample size compared to the full solution. We further report the relative speedup compared to FULL. Since the algorithms are randomized, we run them 50 times with different random seeds and compute sample averages with corresponding 95% confidence intervals based on the standard error of the mean. All experiments were run on an Intel Xeon machine with 36× 2.3GHz processors and 1.5TB memory.

**Data sets.** We consider the $k$-means clustering problem on four different data sets for both $k = 100$ and $k = 500$:
(1) KDD — 145'751 samples with 74 features measuring the match between a protein and a native sequence [18].
(2) CSN — 7GB of cellphone accelerometer data processed into 80'000 observations and 17 features [8].
(3) SONG — 90 features from 515'345 songs of the Million Song datasets used for predicting the year of songs [6].
(4) RNA — 8 features from 488'565 RNA input sequence pairs [26].

**Discussion of the results.** Figure 1 shows the relative error of UNIFORM, LWCS and CS for different subsample sizes $m$. The approximation error decreases for all methods as the sample size is decreased. CS provides substantial improvements compared to UNIFORM on all the data sets considered and for both $k = 100$ and $k = 500$. LWCS retains most of these improvements as it performs roughly as good as CS (on KDD and SONG) or slightly worse (on CSN and RNA). Figure 2 displays the relative error in relation to the time required to construct the coreset and to then solve on the coreset. As lightweight coresets (LWCS) are much cheaper to construct, they strongly outperform the "traditional" construction (CS) across all data sets. Furthermore, they also produce better solutions than UNIFORM except for the smaller sample sizes on SONG and RNA.

The practical impact is evident from Table 1. For KDD with $k = 100$ and $m = 1000$, UNIFORM leads to a speedup of 2244× compared to solving on the full data set but also incurs a high relative error of 195.1%. CS reduces the relative error to 16.0% but only obtains a speedup of 125×. Lightweight coresets capture the best of both worlds — a speedup of 828× at a relative error of only 18.5%. In absolute terms, one may compute 100 cluster centers on a 145 751 point data set in 0.42 seconds compared to 345 seconds if one naively uses the full data set.

## 9 CONCLUSION

We introduced and motivated lightweight coresets — a novel notion of coresets that allows for both multiplicative and additive errors. We proposed a simple and practical algorithm for lightweight coreset construction with corresponding theoretical guarantees on the solution quality. Empirically, the computed coresets match the quality of traditional constructions while they are computed at a fraction of the cost. Furthermore, the approach naturally generalizes to the statistical setting and allows for the construction of substantially smaller summaries for empirical risk minimization than existing state-of-the-art results from learning theory.

## ACKNOWLEDGMENTS

This research was partially supported by SNSF NRP 75 (project number 407540_167212), ERC StG 307036, a Google Ph.D. Fellowship and an IBM Ph.D. Fellowship. This work was done in part while Mario Lucic was at ETH Zurich.

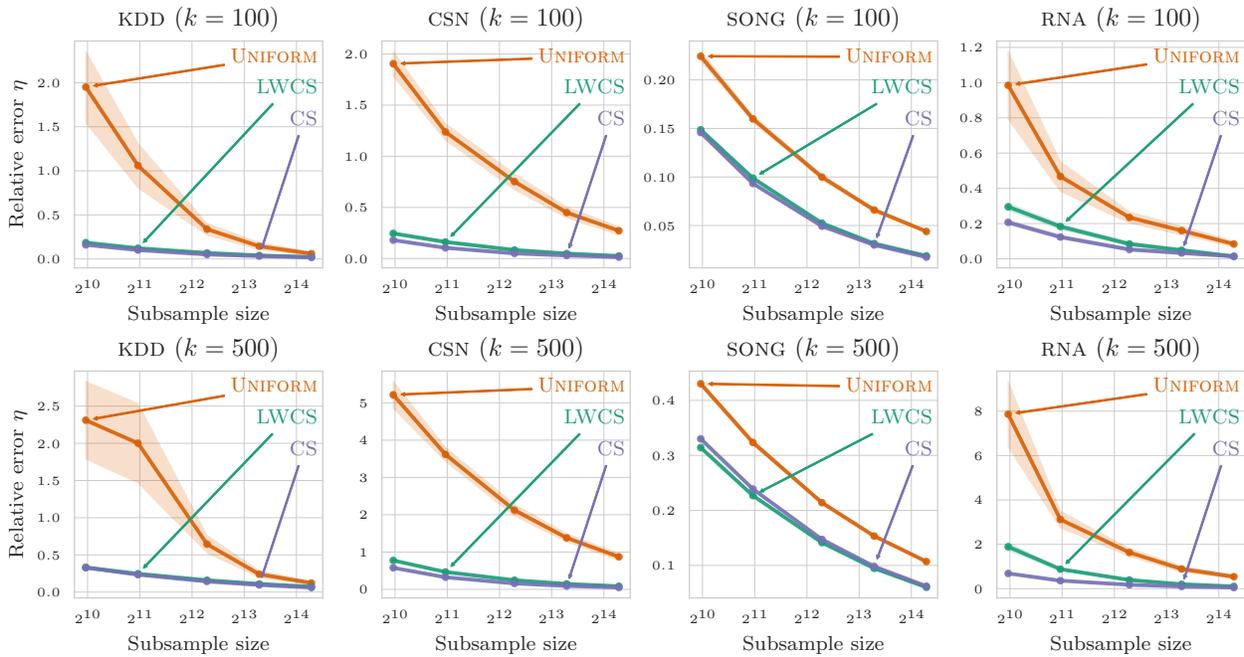

Figure 1: Relative error in relation to subsample size for UNIFORM, LWCS and CS. LWCS captures most of the benefits of CS over UNIFORM. Results are averaged across 50 iterations and shaded areas denote 95% confidence intervals.

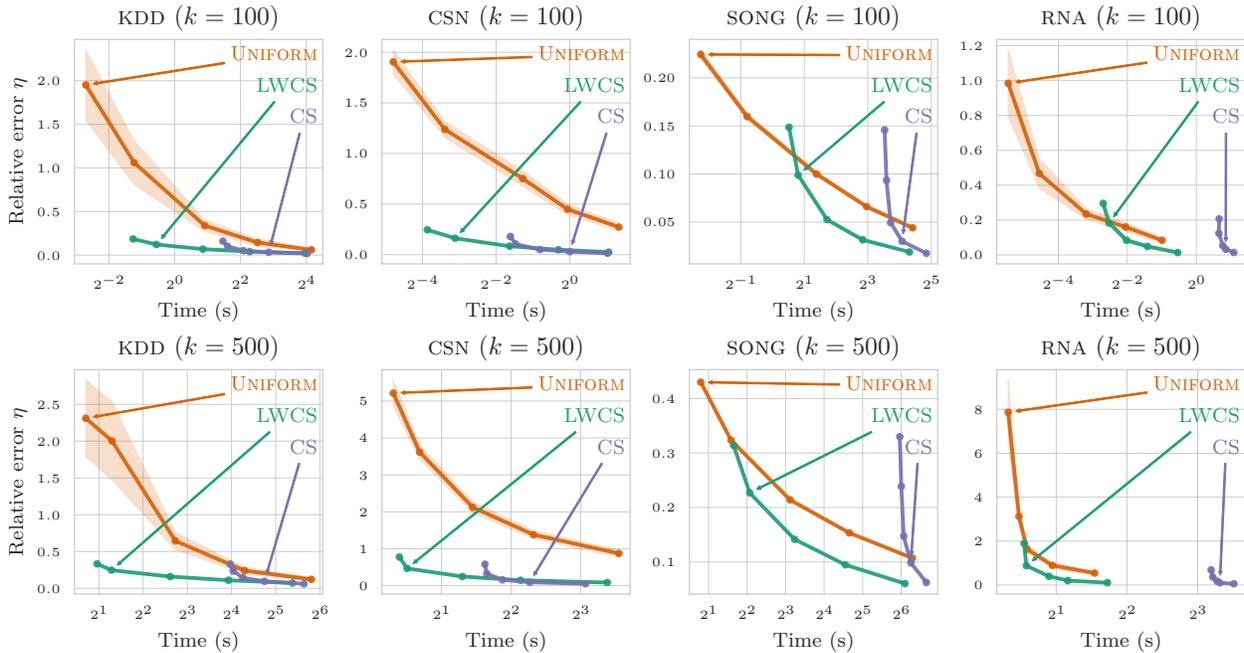

Figure 2: Relative error vs. time for UNIFORM, LWCS and CS. LWCS outperforms CS substantially. Results are averaged across 50 iterations and shaded areas denote 95% confidence intervals. Note that computing the solution on the full data set takes up to $8604$ seconds (SONG, $k = 500$).

Table 1: Relative error and speedup of different methods vs. Full for $k = 100$

| $k$ | Data | Method | Relative error vs. Full | | | Speedup vs. Full | | |
|---|---|---|---|---|---|---|---|---|
| | | | $m = 1000$ | $m = 2000$ | $m = 5000$ | $m = 1000$ | $m = 2000$ | $m = 5000$ |
| 100 | KDD | Uniform | 195.1% ± 20.7 | 105.9% ± 12.9 | 33.8% ± 3.4 | 2244.0× | 809.1× | 183.0× |
| | | LWCS | 18.5% ± 0.2 | 12.1% ± 0.2 | 6.8% ± 0.2 | 828.1× | 506.0× | 190.2× |
| | | CS | 16.0% ± 0.3 | 10.1% ± 0.2 | 5.1% ± 0.1 | 124.8× | 113.3× | 81.1× |
| | CSN | Uniform | 190.5% ± 6.2 | 123.8% ± 4.2 | 75.2% ± 4.0 | 557.7× | 210.9× | 48.7× |
| | | LWCS | 24.6% ± 0.5 | 16.2% ± 0.3 | 8.4% ± 0.3 | 293.5× | 174.0× | 62.2× |
| | | CS | 18.0% ± 0.4 | 10.5% ± 0.3 | 5.0% ± 0.3 | 61.4× | 55.6× | 35.2× |
| | SONG | Uniform | 22.4% ± 0.3 | 16.0% ± 0.2 | 10.0% ± 0.1 | 8027.1× | 2912.1× | 639.8× |
| | | LWCS | 14.9% ± 0.1 | 9.9% ± 0.1 | 5.2% ± 0.0 | 1168.2× | 957.5× | 509.3× |
| | | CS | 14.6% ± 0.1 | 9.3% ± 0.1 | 4.9% ± 0.0 | 144.9× | 139.1× | 127.0× |
| | RNA | Uniform | 98.5% ± 9.8 | 46.8% ± 4.3 | 23.5% ± 1.6 | 773.9× | 414.2× | 161.3× |
| | | LWCS | 29.5% ± 1.0 | 18.3% ± 0.8 | 8.4% ± 0.6 | 114.7× | 101.9× | 71.8× |
| | | CS | 20.7% ± 0.7 | 12.4% ± 0.5 | 5.3% ± 0.5 | 11.2× | 11.2× | 10.4× |
| 500 | KDD | Uniform | 231.0% ± 26.6 | 200.2% ± 27.2 | 64.5% ± 6.1 | 979.8× | 648.8× | 240.6× |
| | | LWCS | 33.1% ± 0.2 | 24.6% ± 0.2 | 15.9% ± 0.1 | 820.0× | 655.5× | 260.7× |
| | | CS | 32.9% ± 0.4 | 23.4% ± 0.2 | 14.3% ± 0.1 | 101.7× | 97.1× | 83.3× |
| | CSN | Uniform | 521.5% ± 18.3 | 361.7% ± 9.7 | 212.3% ± 8.0 | 79.3× | 60.9× | 35.9× |
| | | LWCS | 77.5% ± 0.9 | 46.5% ± 0.4 | 24.5% ± 0.2 | 74.6× | 69.1× | 39.8× |
| | | CS | 57.8% ± 0.8 | 32.4% ± 0.3 | 15.8% ± 0.2 | 31.8× | 31.2× | 26.7× |
| | SONG | Uniform | 43.0% ± 0.2 | 32.3% ± 0.2 | 21.4% ± 0.1 | 4951.6× | 2866.3× | 992.2× |
| | | LWCS | 31.4% ± 0.1 | 22.7% ± 0.1 | 14.2% ± 0.0 | 2740.6× | 2051.8× | 914.1× |
| | | CS | 33.0% ± 0.1 | 23.9% ± 0.1 | 14.7% ± 0.0 | 137.8× | 134.1× | 128.4× |
| | RNA | Uniform | 786.5% ± 75.3 | 312.1% ± 19.3 | 163.4% ± 9.2 | 58.2× | 52.4× | 48.4× |
| | | LWCS | 189.2% ± 8.7 | 88.0% ± 2.7 | 39.5% ± 2.0 | 49.8× | 48.7× | 39.0× |
| | | CS | 68.8% ± 1.0 | 36.4% ± 0.4 | 17.8% ± 0.2 | 8.0× | 7.8× | 7.5× |